\documentclass[sigconf,nonacm]{acmart}

\AtBeginDocument{%
  }

\renewcommand\footnotetextcopyrightpermission[1]{}

\newcommand{\blfootnote}[1]{%
  \begingroup
  \renewcommand\thefootnote{}\footnote{#1}%
  \addtocounter{footnote}{-1}%
  \endgroup
}

\usepackage{algpseudocode}
\usepackage{amsmath}
\usepackage{xspace}
\usepackage{graphicx}
\usepackage{xcolor}
\usepackage{booktabs}
\usepackage{subcaption}
\usepackage{placeins}

\newcommand{\method}{TRACE\xspace}
\newcommand{\decidon}{DECIDON\xspace}
\newcommand{\corpusense}{Corpusense\xspace}
\newcommand{\mezanno}{Mezanno\xspace}

\begin{document}

\title{TRACE: Accountable Agentic Retrieval for Source Discovery in Digital Archives}

\author{Donghan Bian}
\orcid{0009-0007-1684-6026}
\affiliation{%
  \institution{École nationale des chartes -- PSL}
  \department{Centre Jean-Mabillon}
  \city{Paris}
  \country{France}}
\affiliation{%
  \institution{EPITA}
  \department{EPITA Research Laboratory}
  \city{Le Kremlin-Bicêtre}
  \country{France}}

\author{Marie Puren}
\orcid{0000-0001-5452-3913}
\affiliation{%
  \institution{EPITA}
  \department{EPITA Research Laboratory}
  \city{Le Kremlin-Bicêtre}
  \country{France}}
\affiliation{%
  \institution{École nationale des chartes -- PSL}
  \department{Centre Jean-Mabillon}
  \city{Paris}
  \country{France}}

\author{Florian Cafiero}
\orcid{0000-0002-1951-6942}
\affiliation{%
  \institution{EPITA}
  \department{EPITA Research Laboratory}
  \city{Le Kremlin-Bicêtre}
  \country{France}}
\affiliation{%
  \institution{Geneva Graduate Institute}
  \department{Centre for Digital Humanities and Multilateralism}
  \city{Geneva}
  \country{Switzerland}}
\renewcommand{\shortauthors}{Donghan Bian, Marie Puren, and Florian Cafiero}

\begin{abstract}
Historical archives pose a difficult retrieval problem for retrieval-augmented generation systems: documents are OCR-degraded, heterogeneous across genres and sources, and require strong source traceability for scholarly and institutional use. We introduce \method, a training-free agentic retrieval framework designed for accountable source discovery over historical corpora. The system was developed in the context of \decidon, an interdisciplinary project on the circulation of political discourse between parliamentary debates and the press during the French Third Republic, involving digitised historical collections and institutional use cases. The prototype is currently deployed internally within the project and accessible to 24 researchers across six partner institutions. We evaluate \method on HistoriQA-ThirdRepublic, a benchmark of 1{,}752 French historical questions over parliamentary debates and newspapers from 1887, with documents derived from Bibliothèque nationale de France digitised collections. \method achieves R@10 = 0.856 and MRR = 0.653, outperforming sparse, dense, graph-based, and agentic RAG baselines, with the largest gains on multi-hop and cross-corpus questions. At approximately \$0.02 per question under the default hosted inference configuration, \method also remains economically feasible for heritage institutions, laboratories or companies that cannot rely on costly local GPU infrastructure. These results suggest that, for large digital libraries and archives, retrieval accountability and corpus-aware agent design can provide a practical alternative to heavier training-based or graph-construction approaches.
\end{abstract}




\maketitle

\blfootnote{This is the authors' version of the work, posted for personal and
non-commercial use. The definitive version was accepted for publication in the
\emph{Proceedings of the 35th ACM International Conference on Information and
Knowledge Management (CIKM '26)}, November 7--11, 2026, Rome, Italy, and is
available at \url{https://doi.org/10.1145/3799682.3840134}.}

%% ===================================================================
\section{Introduction}
\label{sec:intro}

Archival documents have long served as a cornerstone of historical
research~\citep{blouinjr.2011}. With the convergence of computer science and digital
humanities, large volumes of paper-based materials have been digitised, making
large-scale search increasingly feasible. Yet deploying retrieval-augmented generation
(RAG)~\citep{lewis2020a} in historical research settings remains hindered by three
compounding challenges. At the \emph{technical} level, high OCR error rates introduce
semantic drift during embedding and degrade retrieval
precision~\citep{bourne2026,zhang2025}. At the \emph{query} level, historical questions
routinely span multiple heterogeneous sources, extended temporal scopes, or require
cross-document synthesis rather than single-fact lookup. At the \emph{methodological}
level, most RAG systems optimise for generation quality~\citep{nguyen2025,xia2026a},
whereas historical scholarship imposes strict source traceability, interpretive
transparency, and the possibility of human intervention at every retrieval
step~\citep{puren2026}, requirements with no counterpart in general-purpose RAG
settings. These constraints are also shared by libraries, archives, and institutional
repositories that hold large digitised collections but lack corpus-specific supervision or
local GPU infrastructure.

In response, we propose \method (Training-free Retrieval with Agentic Corpus
Exploration), a lightweight, training-free agentic RAG framework designed for accountable
source discovery over historical corpora. This work is grounded in
\decidon,\footnote{\url{https://anr.fr/Projet-ANR-25-CE38-4063}} a French National
Research Agency project on political discourse circulation between parliamentary debates
and the press during the French Third Republic, involving computer scientists, historians,
and the Bibliothèque nationale de France (BnF). In historical and archival research,
retrieval is not a hidden preprocessing step but a scholarly operation in its own right: a
retrieved source must be discoverable, inspectable, and citable. \method is also intended
to complement \corpusense, the IIIF-based tool developed within the \mezanno project for
transforming digitised serial sources into structured
data~\citep{abadie2026mezanno,mezanno2026corpusense,bnf2026mezanno}. An open-source version of TRACE is also released \footnote{\url{https://github.com/Kepler1908/TRACE}}.

Our contributions are threefold:
\begin{itemize}
  \item We reformulate historical archival retrieval as \emph{source discovery}: an
  applied, corpus-aware document retrieval problem under OCR degradation, heterogeneous
  source structure, and source-traceability constraints, evaluated on retrieval quality
  rather than on generation metrics.
  \item We introduce \method, a training-free agentic retrieval framework combining
  corpus-targeted decomposition, fused warm start, explicit retrieval verdicts with
  deferred re-evaluation, and count-grouped reranking over agent-authored rationales.
  \item We show empirically that this orchestration of mature, off-the-shelf retrieval
  components outperforms purpose-built graph-based and agentic RAG systems on a
  1{,}752-question French historical benchmark, while remaining training-free and
  costing approximately \$0.02 per question.
\end{itemize}

%% ===================================================================
\section{Related Work}
\label{sec:related}

\subsection{From Retrieval Effectiveness to Retrieval Accountability}
Agentic RAG has emerged in two broad forms: training-based methods using reinforcement
learning~\citep{asai2023,li2025a,hui2026a,xia2026a} and training-free methods relying
on framework design~\citep{besrour2025a,nguyen2025,du2026}. Both demonstrate meaningful
retrieval gains, but share a critical limitation: they are evaluated end-to-end on
generation metrics (Exact Match, F1, LLM-judged accuracy), leaving retrieval quality
unexamined and conflating parametric knowledge with grounded evidence
retrieval.\footnote{Self-RAG~\citep{asai2023} and RAGentA~\citep{besrour2025a} do report
recall, but as one metric among several.} Graph-based
RAG~\citep{edge2025a,guo2025,gutierrez2025b,zhuang2025} offers complementary structural
indexing with consistent multi-hop gains, but introduces two constraints relevant here:
graph construction is a form of lossy compression, and entity extraction degrades severely
under OCR noise, causing structural errors that propagate into
retrieval~\citep{zhuang2025}. Historical archival corpora, which combine substantial OCR
noise, heterogeneous sub-corpora, and strict source-traceability requirements, strain the
assumptions of both paradigms simultaneously, motivating a retrieval-first framework that
operates directly on noisy text without intermediate structural extraction.

\subsection{RAG for Historical Corpora in Digital Humanities}
RAG has been increasingly explored in digital humanities for querying large,
heterogeneous, and OCR-noisy archival collections spanning newspaper
archives~\citep{tran_retrieval_2025,mudet_hybrid_2025}, parliamentary
debates~\citep{perez2025}, classical texts~\citep{fan2025a}, administrative
records~\citep{lee2025}, and personal diary archives~\citep{zhou2025,lin2025}. A
recurring finding is that corpus-specific properties, such as OCR artifacts, archaic vocabulary,
and temporal structure, consistently dominate retrieval performance: named-entity
injection at reranking partially mitigates OCR errors~\citep{tran_retrieval_2025};
date-based metadata filtering substantially improves precision on chronologically dense
collections~\citep{lee2025}; and dense retrieval systematically fails on historical
vocabulary, with sparse retrieval and reranking outperforming embedding-based
approaches~\citep{yu2025a}.

Alongside retrieval performance, a parallel concern has emerged around source
traceability and human oversight~\citep{zhou2025,lee2025,perez2025}: in historical
scholarship, identifying relevant sources is a scholarly objective in its own right, and
retrieval must preserve an inspectable, citable path from query to evidence. These
studies motivate a retrieval-first evaluation perspective in which source discovery, not
only answer generation, becomes the object of measurement.

%% ===================================================================
\section{Methodology}
\label{sec:method}

\subsection{Problem Setting and Architecture Overview}
\label{sec:method:setting}

Let $\mathcal{D} = \mathcal{D}_1 \cup \cdots \cup \mathcal{D}_K$ be a heterogeneous archival corpus partitioned into $K$ sub-corpora that differ in genre, vocabulary, and document quality.
Given a question $q$ in the language of the corpus, the system must return an ordered list of document identifiers.
The gold set $G_q \subseteq \mathcal{D}$ may span multiple sub-corpora and is not recoverable from the surface form of $q$.
We evaluate with recall@$k$ as the primary metric.

\begin{figure*}[t]
\centering
\includegraphics[width=1\textwidth]{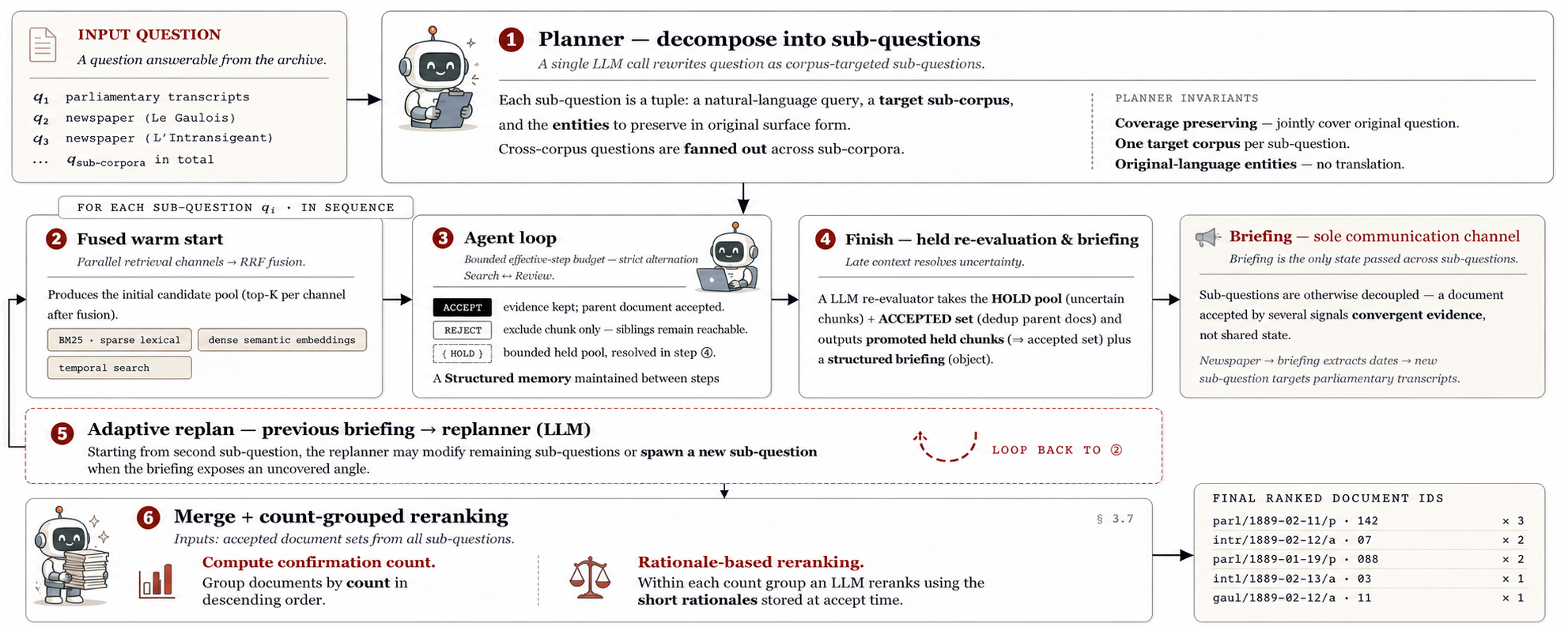} 
\caption{Overview of the \method pipeline.}
\Description{Pipeline diagram showing a user question decomposed into sub-questions, followed by fused retrieval, an agentic search-and-review loop, held re-evaluation, briefing and replanning, and final document merging and reranking.}
\label{fig:arch}
\end{figure*}

\method addresses the challenges identified above through a six-stage pipeline, as shown in Figure~\ref{fig:arch}.

\subsection{Planner: Corpus-Targeted Decomposition}
\label{sec:method:planner}

The planner is a single LLM call that decomposes $q$ into a set of $M$ sub-questions $\{s_1, \ldots, s_M\}$, where $M \leq M_{\max}$.
Each $s_i$ is a tuple $(\text{text}_i, c_i, \text{entities}_i)$ where $c_i \in \mathcal{C}$ identifies the target sub-corpus.
The design follows the broader intuition of plan-and-execute and decomposition-based agentic retrieval \citep{wang2023,asai2023, ammann2025}.
The prompt enforces three invariants:
(i)~the sub-questions jointly cover the meaning of $q$ without loss or addition;
(ii)~each $s_i$ targets exactly one sub-corpus, so cross-corpus questions are fanned out into per-corpus sub-questions;
(iii)~entity names retain their original-language surface form to avoid OCR mismatch from translation.

Corpus targeting guides Phase~1 candidate allocation but does not restrict the agent loop: the agent may search across corpus boundaries when evidence suggests cross-corpus connections (\S\ref{sec:method:phase2}).

\subsection{Phase~1: Fused Warm Start}
\label{sec:method:phase1}

For each sub-question $s_i$, Phase~1 produces an initial candidate pool by fusing complementary retrieval channels via reciprocal-rank fusion (RRF, $k{=}60$) \citep{cormack2009}. Each channel independently scores chunks and \emph{max-pools} per parent document: a document with $N$ chunks contributes exactly once per channel---its best chunk score---preventing any single long document from dominating the pool through sheer volume of passages. This hybrid design follows recent findings that sparse, dense, metadata-aware, and reranking strategies often compensate for different failure modes in historical corpora \citep{lee2025,yu2025a}.

\begin{description}
  \item[Sparse lexical (BM25)]\citep{robertson2009} retrieves over tokenised chunks and is effective for proper nouns and domain terms that survive OCR.
  \item[Dense semantic] uses cosine similarity over pre-computed chunk embeddings, capturing paraphrases and conceptual matches that lexical search misses.
  \item[Temporal] uses date-based retrieval with exponential decay scoring by temporal distance from the target date, activating when a sub-question carries an explicitly resolvable date \citep{lee2025}.
\end{description}

The two principal channels are strongly complementary rather than redundant. Replaying the executed sub-questions through each channel independently, the mean Jaccard overlap between the BM25 and dense top-20 pools is only $0.19$, and $28.8\%$ of the gold documents recovered by the fusion are reachable through exactly one of the two. Fusing them retrieves $78.9\%$ of gold documents against $70.1\%$ for the better single channel. This provides implicit robustness to OCR noise without explicit preprocessing: BM25 matches tokens that survive digitisation, while dense retrieval captures semantic paraphrases of corrupted text. In an imbalanced corpus, however, applying a global result cap systematically truncates minority sub-corpus documents. We address this by pre-computing the set of document IDs belonging to the target sub-corpus and passing these as \emph{candidate constraints} to each channel, so the cap applies \emph{within} the target sub-corpus rather than across the full collection.

\subsection{Phase~2: Agent Loop}
\label{sec:method:phase2}

The warm-start pool seeds a bounded agent loop with an \emph{effective step budget}. At each step, the agent either issues a new search query---through sparse, dense, temporal, or exact-substring search---or reviews a pending batch of retrieved chunks. Search and review strictly alternate: after each search returns candidates, the agent must review them before issuing another query, ensuring that each retrieval is immediately exploited and subsequent queries can build on newly discovered evidence. This design is related to ReAct-style alternation between reasoning and action \citep{yao2023}, but adapts it to source discovery by making retrieval verdicts explicit. Every reviewed chunk receives one of three verdicts:

\begin{description}
  \item[\textsc{accept}] The chunk provides evidence toward answering $q$. The agent records a short free-text rationale (${\leq}120$ characters, reused at reranking time), and the parent document is deduplicated from subsequent retrieval.
  \item[\textsc{reject}] The chunk is off-topic. Only the chunk---not the parent document---is excluded from future retrieval, preserving access to sibling passages of the same long document.
  \item[\textsc{hold}] The chunk is possibly relevant but uncertain given current context. It enters a bounded \emph{held pool} for deferred re-evaluation (\S\ref{sec:method:finish}).
\end{description}

The agent is always presented with both $q$ and $s_i$: the sub-question focuses the search direction, but relevance is defined with respect to the original question, ensuring that cross-cutting evidence is not discarded by an overly narrow interpretation of the current sub-question.

Temporal search receives special treatment: the agent's specified date window is silently expanded by one day on each boundary, compensating for indeterminacy in date-boundary decisions without exposing the expansion to the agent. Exposing it would risk anchoring subsequent temporal reasoning on artificially widened boundaries. Results from date expansion are conservatively auto-held rather than accepted, entering the deferred pool for later re-evaluation.

To prevent circular exploration, the agent maintains a five-slot structured memory: prior findings, question analysis, evidence gathered, gaps remaining, and next steps. This memory is updated by \emph{replacement} at every step. Replacement semantics force the agent to synthesise rather than accumulate, keeping the memory compact and current. Alongside this, the system tracks an \emph{effective step counter} that increments only on productive actions. Duplicate queries, parse failures, and empty results are skipped with informative feedback, preserving the step budget for genuinely new exploration; two consecutive identical search attempts trigger forced termination to prevent degenerate loops.

Each agent step is a single-shot LLM call, receiving a system prompt and a constructed user prompt in JSON mode. The agent retains no memory of prior API calls; all state is reconstructed from the structured memory and a windowed action log containing the last eight entries. The output is a JSON object (\texttt{thought}, \texttt{action}, \texttt{args}, \texttt{memory\_update}) parsed without tool-calling APIs. This design eliminates context pollution from accumulated conversation history, makes calls stateless and parallelisable across sub-questions, and renders the system provider-agnostic: any LLM supporting JSON-mode output is sufficient, with no dependency on proprietary function-calling interfaces.

\subsection{Finish, Briefing, and Replan}
\label{sec:method:finish}

When the agent terminates, a \emph{held re-evaluation} call presents the full accepted set, the stage memory, and summaries of all held chunks to the LLM, which promotes any held chunk now supported by the accumulated evidence. This is the payoff of the three-way verdict: early uncertainty is resolved against late context rather than collapsed into a premature binary decision at first encounter.

A \emph{briefing} is then synthesised as a structured object containing a narrative summary, confirmed facts, confirmed dates, confirmed entities, remaining gaps, and search hints. It serves as the sole communication channel between sub-questions. This enables later sub-questions to start from the evidence frontier rather than searching blind. Before each non-initial sub-question $s_i$ ($i{>}1$), a replanner receives the previous briefing and the remaining plan, and may modify remaining sub-questions or spawn new ones when the briefing reveals gaps not covered by the original decomposition. The total number of sub-questions is capped at $M_{\max}$ to bound cost. This transforms the retrieval plan from a static script into an adaptive investigation while keeping sub-question processing auditable.

\subsection{Merge and Count-Grouped Reranking}
\label{sec:method:merge}

After all sub-questions complete, their accepted sets $\{A_1, \ldots, A_M\}$ are merged. Because sub-questions are structurally decoupled and share no accepted sets during their respective agent loops, a document accepted by multiple sub-questions represents convergent evidence from separate investigative paths rather than confirmation bias from shared state. For each document $d \in \bigcup_i A_i$ we compute a \emph{confirmation count} $\mathrm{cnt}(d) = |\{i : d \in A_i\}|$, and the final ranking proceeds in two stages. First, documents are partitioned by $\mathrm{cnt}(d)$ and groups are placed in decreasing count order. Within each group of size ${\geq}\,2$, an LLM ranks the documents using the concatenation of the agent's own acceptance rationales---the ${\leq}120$-character justifications recorded during the agent loop---rather than re-reading full document text. Singleton groups pass through without an LLM call, so the reranker's cost is proportional to the number of ties rather than the total candidate count.

%% ===================================================================
\section{Experimental Setup}
\label{sec:exp}

\subsection{Corpus and Questions}
\label{sec:exp:dataset}

We use HistoriQA-ThirdRepublic \citep{pellet2026} as our evaluation dataset. It consists of a text corpus and a question-answer dataset built around the same type of historical materials that motivate the \decidon use case: parliamentary debates and newspapers from the French Third Republic. The benchmark therefore serves both as an evaluation resource and as a controlled proxy for the broader applied problem of accountable source discovery over BnF-derived digitised collections.

The corpus consists of 3{,}386 French-language documents from the French Third Republic period (1887), drawn from three heterogeneous sub-corpora: parliamentary debate transcripts from the \emph{Journal Officiel de la République française} and two Parisian daily newspapers (\emph{Le Gaulois}, \emph{L'Intransigeant}) representing distinct editorial positions on the political events of the period. The documents are derived from OCR-digitised historical collections held by the BnF, reflecting the type of large-scale digitised material encountered in national library and archival infrastructures. They contain substantial OCR errors, irregular layouts, and genre-specific vocabulary, making the corpus a realistic testbed for accountable retrieval in digital heritage settings. Documents are split into 5{,}664 sentence-boundary chunks with a maximum of 1{,}000 whitespace tokens and no overlap.

The evaluation set comprises 1{,}752 questions with gold document sets, constructed following the methodology of \citet{pellet2026} in collaboration with a domain historian, who validated the factual and contextual coherence of the generated questions. Questions span four types reflecting patterns of historical inquiry: single-hop (\textbf{SH}, $n{=}889$), where each question targets a single source document; multi-hop generic (\textbf{MH-Generic}, $n{=}529$), requiring cross-source synthesis across parliamentary and press sources; bridge-entity (\textbf{MH-BridgeEnt}, $n{=}142$), where a shared named entity connects documents from different sub-corpora; and comparative (\textbf{MH-Comp}, $n{=}192$), contrasting viewpoints across heterogeneous sources. Gold sets may span multiple sub-corpora, making corpus-aware retrieval essential. Prior evaluation on this corpus demonstrates the severity of the retrieval challenge: standard dense retrieval achieves a Recall@3 of only 14.3\% on cross-newspaper queries and 47.8\% on cross-domain queries \citep{pellet2026}.

\subsection{Baselines and Metrics}
\label{sec:exp:setup}

We conduct three sets of experiments. \textbf{(1) Main results} evaluate \method under two backbone configurations: DeepSeek-V3.2 \citep{deepseek-ai2025a}, the earlier model used during system development, and DeepSeek-V4-Flash \citep{deepsee-v4}, the configuration adopted for the remaining experiments. \textbf{(2) Baseline comparison} compares \method against six systems across three retrieval paradigms: basic retrieval (BM25 and dense retrieval), graph-based RAG (HippoRAG~2 \citep{gutierrez2025b} and LinearRAG \citep{zhuang2025}), and agentic RAG (A-RAG \citep{du2026} and MA-RAG \citep{nguyen2025}). These baselines are selected for the distinctiveness of their design or strong reported performance on multi-hop retrieval benchmarks. \textbf{(3) Cumulative ablation} progressively adds components to a baseline agent loop to quantify the contribution of each mechanism. Experiments (2) and (3) use DeepSeek-V4-Flash and Qwen3-Embedding-8B \citep{zhang2025a} throughout.

All baseline systems were adapted to operate on the French corpus. Adaptations were limited to what correctness required: English prompts and extraction templates were translated, and the SpaCy \citep{honnibal_spacy_2020} pipeline used for entity extraction in the graph-based systems was replaced with its French equivalent. All systems consume an identical corpus conversion---the same documents, the same sentence-boundary chunking at 1{,}000 whitespace tokens, and the same document identifiers---and share the embedding model and LLM backbone wherever applicable, with retrieval depth fixed across systems. Differences therefore reflect retrieval architecture rather than corpus preprocessing or model choice. The conversion and adaptation scripts are released alongside the system.

For a question with gold set $G_q$ and ranked result list $\pi$, we report:
\[
  \text{R@}k = \frac{|G_q \cap \pi_{1:k}|}{|G_q|}, \qquad
  \mathrm{MRR}(q) = \frac{1}{|G_q|} \sum_{g \in G_q} \frac{1}{\mathrm{rank}_\pi(g)},
\]
where $\mathrm{rank}_\pi(g) = \infty$ if $g \notin \pi$. This multi-gold MRR averages the reciprocal rank over \emph{all} gold documents, penalising systems that surface only the easiest gold document while burying the rest. For agentic systems, we additionally report precision $P_\text{acc} = |A_q \cap G_q|\,/\,|A_q|$ and recall $R_\text{acc} = |A_q \cap G_q|\,/\,|G_q|$ over the accepted set $A_q$, measuring the quality of accept/reject decisions independently of the final ranking. In the main experiments, we also report the average number of accepted documents for each question type.

%% ===================================================================
\section{Results}
\label{sec:results}

\subsection{Main Results}
\label{sec:results:main}

Table~\ref{tab:main} reports per-type retrieval performance on the full 1{,}752-question evaluation set.
The two backbone models exhibit a consistent precision-recall trade-off: DeepSeek-V3.2 achieves higher overall recall by accepting more documents on average, while DeepSeek-V4-Flash yields substantially higher $P_\text{acc}$ with a more selective acceptance strategy.
Both models show the same difficulty gradient across question types: single-hop questions are near-ceiling, multi-hop generic questions maintain high recall but lower MRR, and bridge-entity and comparative questions remain hardest as they require cross-corpus evidence chains.
Across all types, $R_\text{acc}$ closely tracks R@10, indicating that the performance bottleneck lies in the agent loop rather than in the final reranking step. Attributing every missed gold document in the V4-Flash run to a cause confirms this: $82.6\%$ of misses occur inside the loop and only $17.4\%$ are documents that were accepted but ranked below rank ten. Within the loop, documents that were surfaced and then judged irrelevant ($36.8\%$ of misses) slightly outnumber those never retrieved by any channel ($36.5\%$), with a further $9.3\%$ held but never promoted. Relevance judgment, not retrieval coverage, is thus the single largest source of error.

\begin{table*}[t]
\caption{Main retrieval results on the full evaluation set (N=1{,}752). \#RET is the mean number of accepted documents.}
\label{tab:main}
\begin{tabular}{lccccccr}
\toprule
\textbf{Type} & \textbf{R@3} & \textbf{R@5} & \textbf{R@10} & \textbf{MRR} & \textbf{P\textsubscript{acc}} & \textbf{R\textsubscript{acc}} & \textbf{\#RET} \\
\midrule
\multicolumn{8}{l}{\textit{DeepSeek-V3.2 (0324)}} \\
SH                & 0.872 & 0.904 & 0.922 & 0.797 & 0.246 & 0.933 & 8.1 \\
MH -- Generic     & 0.684 & 0.808 & 0.892 & 0.532 & 0.250 & 0.932 & 11.0 \\
MH -- BridgeEnt   & 0.644 & 0.761 & 0.820 & 0.508 & 0.276 & 0.838 & 8.5 \\
MH -- Comparative & 0.484 & 0.596 & 0.727 & 0.383 & 0.140 & 0.846 & 21.1 \\
All               & 0.754 & \textbf{0.830} & \textbf{0.884} & 0.648 & 0.238 & \textbf{0.915} & 10.4 \\
\midrule
\multicolumn{8}{l}{\textit{DeepSeek-V4-Flash}} \\
SH                & 0.852 & 0.881 & 0.893 & 0.793 & 0.452 & 0.902 & 4.6 \\
MH -- Generic     & 0.739 & 0.832 & 0.889 & 0.560 & 0.349 & 0.906 & 8.5 \\
MH -- BridgeEnt   & 0.648 & 0.725 & 0.750 & 0.472 & 0.348 & 0.754 & 5.7 \\
MH -- Comparative & 0.500 & 0.578 & 0.674 & 0.396 & 0.275 & 0.789 & 14.9 \\
All               & \textbf{0.763} & 0.820 & 0.856 & \textbf{0.653} & \textbf{0.393} & 0.879 & \textbf{7.0} \\
\bottomrule
\end{tabular}
\end{table*}

The more conservative acceptance behaviour of DeepSeek-V4-Flash should be interpreted as a different operating point rather than a simple degradation: the agent declines to accept marginally relevant documents, improving accepted-set precision while modestly reducing accepted-set recall. This matters in digital-archive workflows, where every accepted document may become part of a human reading queue. A configuration that returns a smaller, more precise accepted set can therefore be preferable even when raw recall is slightly lower.

In our hosted inference setup, DeepSeek-V4-Flash consumed a total of 177.6\,M input tokens and 39.7\,M output tokens, costing approximately \$36 at the API prices used for the experiment. DeepSeek-V3.2 cost approximately \$93, or 158\% more, driven by higher per-token pricing and 1.8$\times$ greater input consumption from deeper agent exploration. At approximately \$0.02 per question, the V4-Flash configuration makes large-scale corpus exploration economically feasible for humanities laboratories operating under tight resource constraints, while remaining well below the practical cost of maintaining equivalent local GPU deployment for occasional or project-based use. Given these deployment advantages, we adopt DeepSeek-V4-Flash as the backbone for practical use.

We also quickly tested our system's latency. On a 200-question timed sample of target dataset, \method issues on average $33.8$ LLM calls per question, totalling $169$\,s of model time. Latency is almost entirely provider-bound: $97.8\%$ of per-question wall-clock is spent awaiting the hosted model, leaving roughly four seconds of framework computation, so responsiveness tracks the serving backend. Because sub-questions and questions are processed independently, serial timing is a worst case; running eight questions concurrently, the same workload completes in $22$\,s per question. Source discovery is therefore an asynchronous operation at archival scale, which matches how the prototype is used in practice, where researchers issue queries continuously or submit a prepared question set.

\subsection{Comparison with Baselines}
\label{sec:results:baseline-comparison}

Table~\ref{tab:baselines} compares \method against representative baselines across three retrieval paradigms.

On basic retrieval, dense retrieval performs well on single-hop questions but degrades substantially on multi-hop types, a pattern consistent with what \citet{pellet2026} observed on this corpus. Sparse retrieval (BM25) is competitive on single-hop questions but performs poorly on comparative questions, where relevant evidence is dispersed across heterogeneous sources and lexical overlap between the query and the evidence may be limited.

Graph-based systems yield a more surprising result: both HippoRAG~2 and LinearRAG perform at levels comparable to dense retrieval across most question types, without recovering their reported multi-hop advantage. Both systems return largely what dense retrieval already returns: $7.1$ and $6.7$ of their top-ten documents respectively, against $3.0$ for BM25. The documents they add are a wash: they gain $98$ and $113$ gold documents over dense while losing $86$ and $132$, a net of $+12$ and $-19$ out of roughly $1{,}900$. The graph layer therefore perturbs the dense ranking without contributing retrieval signal. 

\method achieves consistent and substantial gains across all question types, with the margin widening on multi-hop questions that require cross-corpus evidence chains.

\begin{table*}[t]
\caption{Baseline comparison on HistoriQA-ThirdRepublic. Panel (a) reports R@10 and MRR for all systems. Panel (b) reports accepted-set precision and recall for agentic systems.}
\label{tab:baselines}

\begin{minipage}[t]{\textwidth}
\centering
\subcaption{R@10 and MRR per question type (all systems)}
\resizebox{0.7\textwidth}{!}{%
\begin{tabular}{lcccccccccc}
\toprule
\textbf{System}
  & \multicolumn{2}{c}{\textbf{SH}}
  & \multicolumn{2}{c}{\textbf{MH -- Generic}}
  & \multicolumn{2}{c}{\textbf{MH -- BridgeEnt}}
  & \multicolumn{2}{c}{\textbf{MH -- Comp}}
  & \multicolumn{2}{c}{\textbf{All}} \\
\cmidrule(lr){2-3}\cmidrule(lr){4-5}\cmidrule(lr){6-7}\cmidrule(lr){8-9}\cmidrule(lr){10-11}
 & R@10 & MRR & R@10 & MRR & R@10 & MRR & R@10 & MRR & R@10 & MRR \\
\midrule
\multicolumn{11}{l}{\textit{Basic retrieval}} \\
BM25             & 0.774 & 0.606 & 0.596 & 0.381 & 0.623 & 0.408 & 0.180 & 0.098 & 0.643 & 0.467 \\
Dense retrieval  & 0.879 & 0.722 & 0.665 & 0.400 & 0.665 & 0.413 & 0.302 & 0.129 & 0.734 & 0.535 \\
\midrule
\multicolumn{11}{l}{\textit{Graph-based RAG}} \\
HippoRAG 2       & 0.879 & 0.581 & 0.669 & 0.371 & 0.665 & 0.392 & 0.323 & 0.130 & 0.737 & 0.453 \\
LinearRAG        & 0.875 & 0.696 & 0.656 & 0.377 & 0.673 & 0.413 & 0.281 & 0.119 & 0.727 & 0.513 \\
\midrule
\multicolumn{11}{l}{\textit{Agentic RAG}} \\
A-RAG            & 0.439 & 0.248 & 0.385 & 0.209 & 0.430 & 0.225 & 0.107 & 0.037 & 0.385 & 0.211 \\
MA-RAG           & 0.636 & 0.422 & 0.603 & 0.325 & 0.630 & 0.377 & 0.273 & 0.121 & 0.586 & 0.356 \\
\method~(ours)   & \textbf{0.893} & \textbf{0.793} & \textbf{0.889} & \textbf{0.560} & \textbf{0.750} & \textbf{0.472} & \textbf{0.674} & \textbf{0.396} & \textbf{0.856} & \textbf{0.653} \\
\bottomrule
\end{tabular}%
}
\end{minipage}

\vspace{1em}

\begin{minipage}[t]{\textwidth}
\centering
\subcaption{Agent decision quality ($P_\text{acc}$ / $R_\text{acc}$) per question type (agentic systems only)}
\resizebox{0.7\textwidth}{!}{%
\begin{tabular}{lcccccccccc}
\toprule
\textbf{System}
  & \multicolumn{2}{c}{\textbf{SH}}
  & \multicolumn{2}{c}{\textbf{MH -- Generic}}
  & \multicolumn{2}{c}{\textbf{MH -- BridgeEnt}}
  & \multicolumn{2}{c}{\textbf{MH -- Comp}}
  & \multicolumn{2}{c}{\textbf{All}} \\
\cmidrule(lr){2-3}\cmidrule(lr){4-5}\cmidrule(lr){6-7}\cmidrule(lr){8-9}\cmidrule(lr){10-11}
 & $P_\text{acc}$ & $R_\text{acc}$ & $P_\text{acc}$ & $R_\text{acc}$ & $P_\text{acc}$ & $R_\text{acc}$ & $P_\text{acc}$ & $R_\text{acc}$ & $P_\text{acc}$ & $R_\text{acc}$ \\
\midrule
A-RAG            & 0.049 & 0.606 & 0.074 & 0.489 & 0.084 & 0.514 & 0.016 & 0.185 & 0.056 & 0.517 \\
MA-RAG           & 0.056 & 0.786 & 0.127 & 0.682 & 0.150 & 0.701 & 0.061 & 0.323 & 0.085 & 0.697 \\
\method~(ours)   & \textbf{0.452} & \textbf{0.902} & \textbf{0.349} & \textbf{0.906} & \textbf{0.348} & \textbf{0.754} & \textbf{0.275} & \textbf{0.789} & \textbf{0.393} & \textbf{0.879} \\
\bottomrule
\end{tabular}%
}
\end{minipage}

\vspace{0.5em}
\end{table*}

The two agentic baselines, A-RAG and MA-RAG, do not produce explicit ranked document lists, instead generating answers from their full retrieval trajectory. To place them on the same footing as the other systems, we rank each trajectory by \emph{order of discovery} and evaluate the resulting list. This is a generous protocol: it credits the baselines with any relevant document encountered at any point in their trajectory, regardless of whether it was used in the generated answer, and their trajectories are long enough for R@10 to be meaningful (median $16$ and $14$ documents respectively). It is also asymmetric in their disfavour in one respect, since discovery order is not a relevance ranking, whereas \method returns a deliberately ranked list. Panel (b) reports the same accepted sets under precision and recall, independently of ordering. Under both views \method achieves substantially higher precision and recall simultaneously, indicating that the advantage stems from the quality of individual retrieval decisions rather than from a larger candidate pool.

The purpose of this comparison is not simply to establish that \method outperforms existing systems. More importantly, it surfaces a structural blind spot in how RAG systems are typically evaluated: generation-centric metrics obscure the quality of the underlying retrieval process, and systems optimised for open-domain QA do not naturally transfer to domain-specific deployment. Effective retrieval in historical research requires targeted design choices.

\subsection{Cumulative Ablation Study}
\label{sec:results:ablation}

Table~\ref{tab:ablation} reports a cumulative ablation in which components are added one at a time to a baseline agent loop. Because the mechanisms in \method are designed to interact---decomposition produces the sub-questions that memory and replanning operate on, while the warm start seeds the pool that the agent loop refines---we adopt a cumulative rather than leave-one-out design, which better reflects the system's intended operating mode.

\begin{table*}[t]
\caption{Cumulative ablation on the full evaluation set (N=1{,}752). Each level adds one component to the previous level. $\Delta$ is the change in overall R@10 from the previous level.}
\label{tab:ablation}
\resizebox{0.8\textwidth}{!}{%
\begin{tabular}{clcccccccccc}
\toprule
 & & \multicolumn{2}{c}{\textbf{SH}} & \multicolumn{2}{c}{\textbf{MH -- Generic}} & \multicolumn{2}{c}{\textbf{MH -- BridgeEnt}} & \multicolumn{2}{c}{\textbf{MH -- Comp}} & \textbf{All} & \\
\cmidrule(lr){3-4} \cmidrule(lr){5-6} \cmidrule(lr){7-8} \cmidrule(lr){9-10}
\textbf{Lv} & \textbf{Added component} & R@10 & MRR & R@10 & MRR & R@10 & MRR & R@10 & MRR & R@10 & $\Delta$ \\
\midrule
0 & Baseline (agent only)    & 0.813 & 0.713 & 0.647 & 0.435 & 0.352 & 0.247 & 0.466 & 0.284 & 0.688 & --- \\
1 & + Decomposition          & 0.790 & 0.699 & 0.849 & 0.539 & 0.637 & 0.413 & 0.625 & 0.368 & 0.777 & +8.9\% \\
2 & + Memory \& replan       & 0.807 & 0.707 & 0.846 & 0.536 & 0.775 & 0.455 & 0.669 & 0.371 & 0.801 & +2.4\% \\
3 & + Phase~1 warm start     & 0.885 & 0.785 & 0.882 & 0.559 & 0.736 & 0.471 & 0.695 & 0.411 & 0.851 & +5.0\% \\
4 & + Hold \& re-eval (full) & 0.893 & 0.793 & 0.889 & 0.560 & 0.750 & 0.472 & 0.674 & 0.396 & 0.856 & +0.5\% \\
\bottomrule
\end{tabular}%
}
\end{table*}

Sub-question decomposition yields the largest gain (+8.9\%), dramatically improving
multi-hop types while leaving single-hop performance largely unchanged. Memory and
replanning (+2.4\%) primarily benefits bridge-entity questions by enabling evidence to
accumulate across sub-question boundaries. The Phase~1 warm start (+5.0\%) stabilises all
question types by seeding the agent with a high-quality initial candidate pool. The
hold-and-re-evaluation mechanism contributes modestly in aggregate (+0.5\%), but acts as a
conservative safety layer: its value is concentrated in marginal cases where a premature
rejection would permanently foreclose a document whose relevance only becomes clear under
late-accumulated context.

%% ===================================================================
\section{Discussion, Deployment, and Future Work}
\label{sec:conclusion}

The results show that accountable source discovery over noisy historical archives benefits less from heavier indexing or model training than from corpus-aware orchestration. \method combines four design choices that proved particularly important in this setting: decomposition into corpus-targeted sub-questions, multi-channel fused retrieval, explicit accept/reject/hold decisions, and final ranking based on cross-sub-question confirmation. On HistoriQA-ThirdRepublic, this design achieves R@10\,=\,0.856 and MRR\,=\,0.653, outperforming sparse, dense, graph-based, and agentic RAG baselines while remaining training-free.

The \decidon use case also highlights why retrieval should be treated as an applied scholarly task rather than as a hidden preprocessing step for answer generation. In digital-heritage settings, users need to identify relevant sources, understand why they were retrieved, and preserve a traceable path from query to evidence. \method is currently deployed as an internal prototype within the \decidon project and is accessible to 24 researchers from six partner institutions: EPITA, EHESS-CRH, Inria, the Bibliothèque nationale de France, the École nationale des chartes, and LARHRA. The prototype has been accessible to project researchers since March 2026 and has supported several hundred exploratory research queries over parliamentary debates and the press. It remains an early-stage internal research service rather than a production deployment, and we report it as such: a systematic evaluation of adoption, workflow impact, and expert reading burden is deferred to future work.

The prototype supports four historical case studies: the politicisation of nature, the Colonial Party, social and tax legislation, and anti-parliamentarianism during the French Third Republic. In these cases, historians use \method to locate relevant passages, compare documentary contexts, and follow the circulation of actors, arguments, vocabularies, and framings over time. The system remains a research service rather than a public-facing BnF service, but the planned integration with \corpusense would support a broader BnF-facing workflow in which \method helps identify and document relevant sources, while \corpusense supports the downstream transformation of selected IIIF collections or serial documents into structured data \citep{abadie2026mezanno,mezanno2026corpusense,bnf2026mezanno}.

Several limitations remain. First, the planner occasionally collapses an inherently multi-hop query into a single sub-question, preventing the replan mechanism from firing. This is rare---multi-hop questions receive two or more sub-questions in roughly $98\%$ of cases---but consequential when it happens: of the eighteen affected questions, thirteen fail, a failure rate of $72\%$ against a $20\%$ base rate. Second, the benchmark covers a single historical year, 1887; generalisation to other periods, languages, and archival genres remains to be tested, and a second corpus is under construction for this purpose. Third, the remaining pipeline constants---step and sub-question budgets, memory size, and the temporal expansion window---were fixed during development on this corpus rather than tuned per question type, and we would expect their optimal values to differ on a collection with different structure. Forth, some prompt-level assumptions, such as the relation between parliamentary debates and newspapers, are corpus-specific and would require adjustment for other documentary settings. Finally, \method is designed so that every returned document carries a traceable path from query to evidence---the sub-question that sought it, the search that surfaced it, and the rationale recorded when it was accepted---but we evaluate retrieval effectiveness rather than the downstream value of that trace. Whether it measurably improves expert trust or reduces verification effort remains open, and is the natural subject of the user study we plan alongside further deployment. Preliminary local-inference tests on an RTX PRO 6000 workstation also showed that smaller models still struggle with multi-step planning and structured-output stability, suggesting that local deployment will require either larger institution-grade infrastructure or a protocol-optimised version of \method.

Future work will focus on making \method a mixed-initiative source discovery system. Rather than placing historians only at the end of the pipeline, we plan to test lightweight checkpoints where expert users can validate or revise the planner's decomposition, inspect held documents before final rejection, and adjust the final accepted set. Additional deployment work will measure latency, token consumption, source reuse, and expert reading burden across larger samples, enabling a more precise evaluation of the trade-off between retrieval accuracy, computational cost, and human workload.

\section*{GenAI Usage Disclosure}
Generative AI tools were used to assist with language editing and code completion during the preparation of this work. All scientific claims, experimental design choices, analyses, reported results, and final text were produced, reviewed, and validated by the authors, who remain fully accountable for the content of the paper.

\bibliographystyle{ACM-Reference-Format}
\bibliography{sample-base}

\end{document}